\newcommand{\fixedcolumnwidth}{\columnwidth}

\documentclass{easychair}

\usepackage{cite}
\usepackage{amsmath,amssymb,amsfonts}
\usepackage{algorithmic}
\usepackage{graphicx}
\usepackage{textcomp}
\def\BibTeX{{\rm B\kern-.05em{\sc i\kern-.025em b}\kern-.08em
    T\kern-.1667em\lower.7ex\hbox{E}\kern-.125emX}}

\usepackage{subfigure}

\newcommand{\xp}[1]{page~\pageref{#1}}
\newcommand{\xf}[1]{Figure~\ref{#1}}
\newcommand{\xfp}[1]{Figure~\ref{#1}, \xp{#1}}

\newcommand{\gipsy}{{GIPSY\index{GIPSY}}}

\newcommand{\ripe}{{RIPE\index{RIPE}\index{Frameworks!RIPE}}}

\newcommand{\dfg}{{DFG\index{DFG}}}
\newcommand{\gmt}{{GMT\index{GMT}}}

\newcommand{\langfontstyle}[1]{\sc{#1}}

\newcommand{\glu}{{\langfontstyle{GLU}\index{GLU}}}

\newcommand{\gipl}{{\langfontstyle{GIPL}\index{GIPL}}}

\newcommand{\lucid}{{\langfontstyle{Lucid}\index{Lucid}}}
\newcommand{\ilucid}{{\langfontstyle{Indexical Lucid}\index{Indexical Lucid}}}

\newcommand{\flucid}{{\langfontstyle{Forensic Lucid}\index{Forensic Lucid}}}

\newcommand{\etal}{\emph{et al}}

\newcommand{\java}{{\langfontstyle{Java}\index{Java}}}

\newcommand{\tool}[1]{\texttt{#1}\index{Tools!#1}}

\newcommand{\api}[1]{\texttt{#1}\index{API!#1}}

\newcommand{\puredata}{PureData\index{PureData}\index{Tools!PureData}}
\newcommand{\jitter}{Jitter\index{Jitter}\index{Tools!Jitter}}
\newcommand{\maxmsp}{Max/MSP\index{Max/MSP}\index{Tools!Max/MSP}}
\newcommand{\bpel}{BPEL\index{BPEL}\index{BPEL}\index{Web Services!BPEL}}

\newcommand{\marf}[0]{MARF\index{MARF}\index{Frameworks!MARF}\index{Libraries!MARF}}

\newcommand{\opengl}[0]{OpenGL\index{OpenGL}\index{Libraries!OpenGL}\index{API!OpenGL}}

\newcommand{\lucidL}[1]{{$\mathit{Lucid}$}($L$) }

		{}

\def\myvert{\raise 2.27pt \hbox{\vrule depth 0pt height 8pt width 0.2mm}}
\def\myarrow{\hspace*{0.43mm}%
             \raise 2.29pt\hbox{\vrule depth 0pt height 8pt width 0.16mm}%
             \hspace*{-0.32mm}%
             $\longrightarrow$
             \ %
             }

\newcommand
	{\graphviz}
	{Graphviz\index{Graphviz}\index{Tools!Graphviz}}

\newcommand
	{\prism}
	{PRISM\index{PRISM}\index{Tools!PRISM}}

\authorrunning{Mokhov \emph{et al.}}

\begin{document}

\title{Toward Multimodal Interaction in Scalable Visual Digital Evidence Visualization Using Computer Vision Techniques and ISS}
\titlerunning{Scalable Mutimodal HCI using ISS for Digital Evidence Visualization}

\author{%
Serguei A. Mokhov\\
\affiliation{\url{mokhov@encs.concordia.ca}}
\and
Miao Song\\
\affiliation{\url{m_song@encs.concordia.ca}}
\and
Jashanjot Singh\\
\affiliation{\url{s_jashan@encs.concordia.ca}}
\and
{Joey Paquet}\\
\affiliation{\url{paquet@encs.concordia.ca}}
\and
{Mourad Debbabi}\\
\affiliation{\url{debbabi@encs.concordia.ca}}
\and
{Sudhir Mudur}\\
\affiliation{\url{mudur@encs.concordia.ca}}
}

\maketitle

\begin{abstract}
Visualization requirements in {\flucid} have to do with different levels of case knowledge abstraction, 
representation, aggregation, as well as the operational aspects as the final long-term goal of this proposal. 
It encompasses anything from the finer detailed representation of hierarchical contexts to 
{\flucid} programs, to the documented evidence and its management, its linkage to programs, 
to evaluation, and to the management of {\gipsy} software networks.
This includes an ability to arbitrarily 
switch between those views combined with usable multimodal interaction. The purpose is to determine 
how the findings can be applied to {\flucid} and investigation case management. It is also 
natural to want a convenient and usable evidence visualization, its semantic linkage and the reasoning machinery for it.
Thus, we propose a scalable management, visualization, and evaluation of digital evidence using the 
modified interactive 3D documentary system -- Illimitable Space System -- (ISS) to represent, semantically link, and provide a 
usable interface to digital investigators that is navigable via different multimodal interaction 
techniques using Computer Vision techniques including gestures, as well as eye-gaze and audio.
\end{abstract}


\nocite{mokhov-phd-thesis-2013}
\nocite{law-order-criminal-intent-2004}
\nocite{iss-multimodal-installation-sa2014-ws}
\nocite{iss-multimodal-case-study-docu-vsmm2014}
\nocite{intensional-programming-1}
\nocite{iss-v3-appy-hour-siggraph2015}
\nocite{iss-v3-appy-hour-gem2015}
\nocite{iss-v2-poster-siggraphasia2015}

\section{Introduction}

We propose a scalable management, visualization, and evaluation of digital evidence using the 
modified interactive 3D documentary component of the Illimitable Space System (ISS) to represent,
semantically link, and provide a usable interface to digital investigators.
%
%

The cyberforensic analysis is one phase of the cybercrime investigation
where the investigators strive to produce credible inferences based on evidential
information. The source of this information is usually the phases that
precede the analysis such as evidence acquisition and encoding. Also, this
information can come from an esoteric set of resources that involves computers
but is not limited to that seem fit as an evidence by the
investigators~\cite{mokhov-phd-thesis-2013,flucid-dfg-viz-pst2011}.

{\lucid} programs are data-flow programs that can be visually composed
and illustrated as data-flow graphs as well. {\flucid} is one such {\lucid}
dialect that enables investigators to specify and reason about cyberforensic
cases. It represents the context of the evaluation of the evidences' by encoding
them along with witness stories, evidential statements and modeling the crime scene
to cross-validate claims against the model and perform event reconstruction,
potentially within large swaths of digital evidence~\cite{mokhov-phd-thesis-2013,flucid-dfg-viz-pst2011}.


In 2004, Gladyshev~\cite{gladyshev-phd-2004} introduced the first formal approach to cybercrime investigation.
Their approach uses Finite State Automata to describe the digital system as a Finite State Machine
for event reconstruction. However, it has an associated learning curve and quite complex
for investigators without a formal background in computer science.
{\flucid} is designed to explicitly address these drawbacks and aims to be usable,
expressive, sound and complete. 

One of the many goals of {\flucid} is usability via scalable visualization of enormous data
under investigation. Recently, there have been significant improvements in the domain of modern
2D and 3D virtual reality environments, which can be easily navigated via a variety of
different multimodal interaction techniques.
{\flucid} aims to up the ante by providing such usability improvements by leveraging
modern multimodal techniques in virtual and augmented reality space for the investigators instead of writing
a {\flucid} program to navigate seamlessly. A combination of gestures, audio commands,
eye gaze and hardware controllers are potential candidates to provide navigational
and interaction abilities to the investigators. It will enable us to extend the
Lucid DFG programming onto {\flucid} case modeling and specification~\cite{mokhov-phd-thesis-2013,flucid-dfg-viz-pst2011}.


The purpose here is to determine the applicability of these findings to
{\flucid} and investigation case management. It is also natural to want a convenient
and usable evidence visualization, their semantic linkage and the appropriate
hardware for the same. The visualization requirements in the context of
{\flucid} revolve around the different levels of the case knowledge abstraction,
its representation, aggregation, and the operational aspects as the final long-term
goal of this proposal. It encompasses everything from the finer detailed representation
of hierarchical contexts to {\flucid} programs, to the documented evidence and
its management. It also includes its linkage to programs, to evaluation and
to the management of {\gipsy} software-defined networks along with an ability to
arbitrarily switch between those views combined with usable multimodal
interaction~\cite{mokhov-phd-thesis-2013,flucid-dfg-viz-pst2011}.

\section{Related Work}

In the context of data-flow programming languages, there are quite a few 
research works and proposals that revolve around graph-based visualizations. 
The work of Faustini proved in particular a visualization of any {\ilucid} program as a
{\dfg}~\cite{denotational-operational-semantics-dataflow}.

In 1995, Jagannathan defined one of the first graph-based visualizations 
proposals for {\lucid} programs. He defined different graphic intensional and 
extensional models for {\glu} programming~\cite{glu-graphical-models-1995}. 
Further, in 1999 Paquet's doctoral work with multidimensional intensional programs 
extended on it, followed by the visual parallel programming idea of Stankovic, 
Orgun, \emph{et al}.~\cite{paquetThesis}~\cite{visual-parallel-programming-2002}.

In 2004, Ding provided practical implementation of 
Paquet's foundational work within {\gipsy} in the form of 2D
{\dfg}s~\cite{paquetThesis,yimin04}~\cite{yimin04}. Ding provided an automatic bidirectional 
translation of the intensional programs between their textual and graphical 
representations by employing  \tool{lefty}'s GUI ({\graphviz}'s) and \tool{dot}'s
languages~\cite{flucid-dfg-viz-pst2011,graphviz,dot-language}.
 
Mokhov proposed an idea of one such ``3D editor'' within
{\ripe}~\cite{mokhov-mcthesis-book-reprint10} to visualize, control
communication patterns and load balancing in {\gipsy}.
The editor's idea is to render graphs in a 3D space to allow its
users to redistribute demands visually in case of imbalance of
workload among the workers.
It can be thought of as a virtual 3D remote control accompanied by
a miniature expert system which can trigger the planning, 
caching and load balancing algorithms to learn and perform
efficiently every time a related {\gipsy} application is run.

Similarly, several authors put forward their works on visualizing the 
configuration, formal systems and load balancing with corresponding graph 
systems
\cite{%
sim-viz-resource-alloc-control,%
visual-config-representation,%
logical-reasoning-with-diagrams,%
graph-transform-visual-languages,%
diagramatic-formal-system-euclidean}.

These works defined key concepts that are consistent with
{\gipsy}~\cite{flucid-dfg-viz-pst2011} visual mechanisms especially, the General Manager 
Tier ({\gmt})~\cite{ji-yi-mcthesis-2011}. Rabah provided the initial 
configuration management and PoC visualization for {\gipsy} nodes and tiers 
via the above mentioned {\gmt}~\cite{graph-based-gmt}.

In 2012, Tao {\etal}.\ proposed another interesting work of 
relevance on the visual representation of event sequences, reasoning, and 
visualization of EHR data~\cite{event-seq-reasoning-vis-ehr-2012}.
Wang {\etal}.\ put forward a temporal search algorithm for event 
visualization of personal history~\cite{temporal-pattern-search-algo-vis-2012}.
Monroe {\etal}.\ noted the challenges of specifying intervals and absences 
in temporal queries and approach those with the use of a graphical
language \cite{spec-intervals-absence-temporal-vis-2013}. This could be of particular 
use for \emph{no-observations}~\cite{mokhov-phd-thesis-2013} in {\flucid} case.
A recent novel HCI concept of documentary knowledge visual representation and 
gesture- and speech-based interaction in the \emph{Illimitable Space System}
(ISS) was put forward by Song~\cite{msong-phdthesis-2012} in 2012. A multimodal 
case management interaction system was proposed for the German police called 
\emph{Vispol Tangible Interface: An Interactive Scenario Visualization}
\footnote{\url{http://www.youtube.com/watch?v=_2DywsIPNDQ}}.

Building upon the above-mentioned works, we propose to illustrate nested evidence, 
crime scene and the reconstructed event flow after revaluation in the form of a 2D or 3D {\dfg}.
The direct impact is to aid the forensic investigators by providing a scalable visualization, 
management of evidence modeling, encoding by {\flucid}~\cite{%
mokhov-phd-thesis-2013,%
flucid-imf08,%
flucid-printer-case-icdf2c-2011,%
self-forensics-through-case-studies} and subsequently its evaluation by {\gipsy}~\cite{flucid-dfg-viz-pst2011}.

\subsection{Conceptual Visualization Design}

Deriving from the related research work in context to visualization of {\lucid} programs, 
a conceptual example of a 2D {\dfg} that corresponds to a simple {\lucid} program produced by
Paquet~\cite{paquetThesis}. Presently, the rendering of the same is by Ding in 2004~\cite{yimin04}  
within the {\gipsy} environment~\cite{flucid-dfg-viz-pst2011}.


In \xfp{fig:observation-3d} is the
conceptual model of hierarchical nesting of the evidential
observation sequences $os$, their individual observations $o$ (consisting
of the properties being observed $(P,min,max,w,t)$, details of which are discussed
in the referenced related works and in \cite[Chapter 7]{mokhov-phd-thesis-2013}).
These 2D conceptual visualizations are proposed
to be renderable at least in 2D or in 3D via an interactive interface to allow modeling complex
crime scenes and multidimensional evidence on demand. The end result is envisioned to look like
either expanding or ``cutting out'' nodes or complex-type results as
exemplified in \xf{fig:nat42dfg-3d}\footnote{cutout image credit is that of Europa found on Wikipedia
\url{http://en.wikipedia.org/wiki/File:PIA01130_Interior_of_Europa.jpg} from NASA}~\cite{flucid-dfg-viz-pst2011}.


\begin{figure}[htpb]
	\centering
	\includegraphics[width=\fixedcolumnwidth]{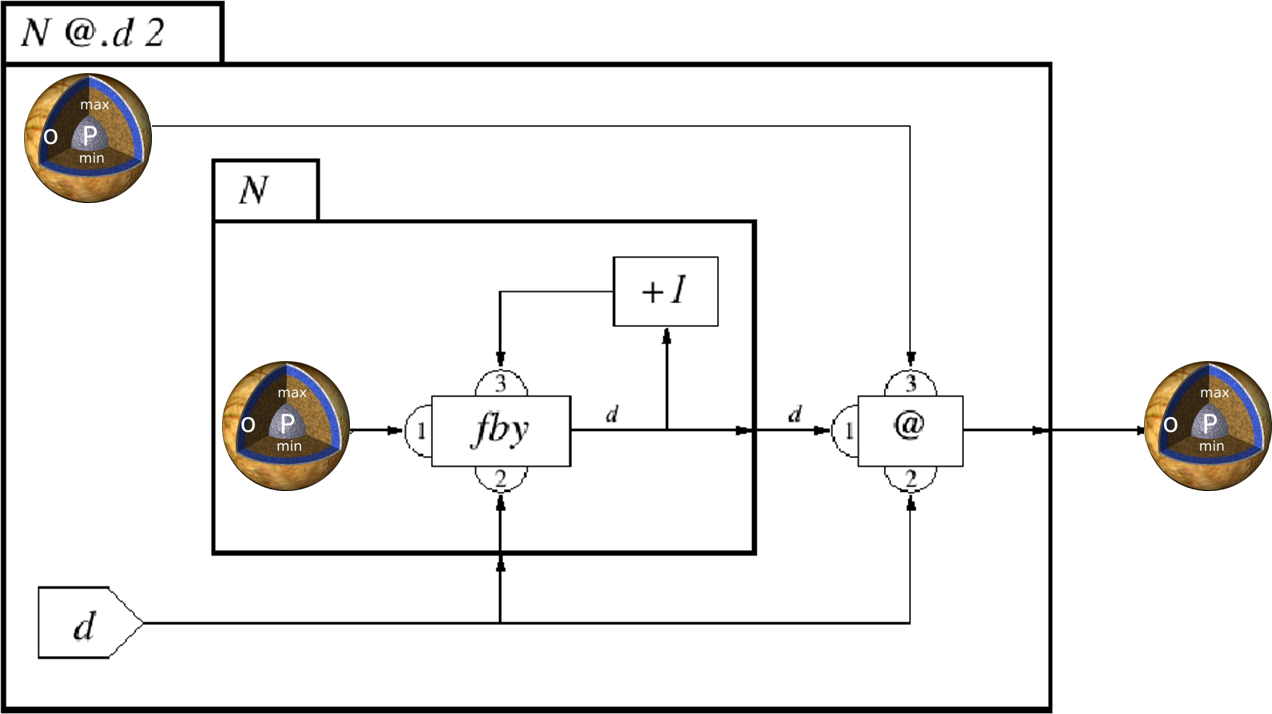}
	\caption{Modified conceptual example of a 2D {\dfg} with 3D elements}
	\label{fig:nat42dfg-3d}
\end{figure}

\begin{figure}[htpb]
	\centering
	\includegraphics[width=.5\columnwidth]{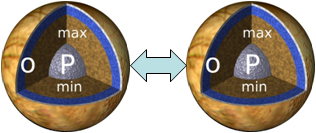}
	\caption{Conceptual example of linked 3D observation nodes}
	\label{fig:observation-3d}
\end{figure}



\section{Multimodal Visual Encoding of {\flucid}-based Evidence}
\label{sect:3d-viz}

Data visualization, not only in the context of cybercrime investigation with 
{\flucid}, but in almost every other domain as well provides numerous 
advantages in terms of deducing inferences, spotting anomalies and 
recognizing patterns.  However, specifically in case of {\flucid} and 
investigating cybercrimes, it provides additional usability~\cite{interaction-design-3ed} enhancements to 
aid investigators to illustrate and define semantic links among the related 
evidence.


Furthermore, the need to visualize forensic cases, 
digital evidence, and related specification components revolve around 
providing usability enhancements to aid the investigators. Additionally, putting the 
program (specification) in 3 dimensions, especially in the modern
and affordable augmented and virtual reality spaces (AR/VR) will help in structuring the program 
along with the case well arranged in a virtual environment with the digital 
evidence enclosed within 3D spheres. Moreover, navigable in depth to whatever 
levels of detail possibly via one of the multimodal interactions, although in 
the given example via clicking, issuing voice commands, gazing, or
gesturing~\cite{flucid-dfg-viz-pst2011}.

In case of event reconstruction, 
in particular, the illustrations and comprehension of operational semantics 
and demand-driven models are much better along keeping in mind their depth 
and complexity. Ding's work provides navigational capabilities from a graph 
to subsequent subgraphs via extending complex nodes to their definitions as 
in \emph{whenever} (\api{wvr}) or \emph{advances upon}
(\api{upon}), their reverse operators, forensic operators, and others~\cite{flucid-dfg-viz-pst2011}
found in \cite[Chapter 7]{mokhov-phd-thesis-2013}.

\subsection{Augmented System Requirements}

In order to realize the envisage of {\dfg} visualization of {\flucid} 
programs and their evaluation by {\gipsy} some immediate considerations are 
discussed below~\cite{flucid-dfg-viz-pst2011}:

\begin{itemize}
\item
Hierarchical evidential statements with deeply nested contexts should be 
visualized~\cite{flucid-dfg-viz-pst2011}.

\item
Intentional-imperative hybrid nodes need to be placed in {\dfg}s
combining fragments of {\lucid} and {\java} programs~\cite{flucid-dfg-viz-pst2011}.
Previous research works by {\gipsy} R\&D did not address the aspects to 
augment the \api{DFGAnalyzer} and \api{DFGGenerator} from Ding's work in some 
fashion to provide support for hybrid {\gipsy} programs. However, to address 
this one can think to add an ``unexpandable'' imperative {\dfg} node to the 
graph, but depth-wise it won't be just enough to click their way through.  
Thus, considering possibilities to make it usable hence expandable recent 
enhancements in {\graphviz} and {\gipsy} can be leveraged to generate {\flucid}
code from the {\dfg} and vice versa~\cite{flucid-dfg-viz-pst2011}.
%


\item
Rabah's work on visualizing load balancing and communication control patterns 
for tasks in Euclidean space may as well be leveraged via the GGMT~\cite{graph-based-gmt}.

\item
The ability to switch among views such as {\dfg}, evidence, and control, 
etc., is required as well.

\item
In flat-screen, touch-screen, augmented reality or projected environments, 3D DFG interactions
are to support click and touch or voice-based call-outs as well as gestures
to link or assemble some of the evidence~\cite{iss-multimodal-case-study-docu-vsmm2014}.

\item
In the virtual reality environment VR controllers and gaze-controlled interactions
are essential in addition to the voice and gesture recognition support~\cite{iss-v3-ar-vr-sa2016}.
\end{itemize}

\subsection{Survey of the Visualization Languages and Tools}

This work focuses on one of the goals of this research, which is to find the 
optimal technique with its formal specifications along with being feasible to 
implement using currently available HCI technologies and a usable one~\cite{flucid-dfg-viz-pst2011}.
%

\subsubsection{{\graphviz}}

Ding's~\cite{yimin04} basic bidirectional translation between {\gipl} and
{\dfg} within {\gipsy} is already a part of the project and exists for {\gipl} 
and {\ilucid}, the two {\lucid} dialect antecedents. Moreover, {\graphviz}
modern version now supports integration with Eclipse~\cite{eclipse}, thus
{\gipsy}'s IDE---{\ripe} (Run-time Interactive Programming Environment)---can be 
an Eclipse-based plug-in as well~\cite{flucid-dfg-viz-pst2011}.

\subsubsection{{\puredata}}

The {\puredata}~\cite{puredata} language by Puckette along with its 
commercial divisions namely ({\jitter}/{\maxmsp}~\cite{jitter}) apply a
{\dfg}-lie programming by graphically placing inlets and outlets of any data 
type connected in the form of so-called ``patches''. These inlets may have 
external implementations and sub-graphs in procedural languages. Originally, 
Puckett's work used signal processing to process electronic music and videos 
in order to produce interactive artistic and performative processes and was
extended beyond that domain. The notion of 
external plug-ins in {\puredata} allows deep visualization of media in {\opengl} which 
in turn enhances the overall aspect of the process. {\puredata} does draw 
influence from {\lucid} as a data flow language as well~\cite{flucid-dfg-viz-pst2011}.

\subsubsection{{\bpel}}

OpenESB IDE provides visual design capabilities to visually illustrate or 
create a {\bpel} (Business Process Execution Language) process along with 
composite applications in the context of Service Oriented Architectures and 
Web Services.~\cite{bpelse,koenig-ws-bpel-2007,ws-bpel-20}. These {\bpel} 
specifications are translatable to an executable web service composition code 
in {\java} language. Not only it provides capabilities in terms of designing 
flows between structures, parallel, asynchronous, sequential processes and 
fault realization, but more importantly, {\bpel} notations have a backing formalism 
modeled upon based on Petri nets.
%
{\bpel} specifications' composite applications actually translate to executable {\java}
web services composition code~\cite{flucid-dfg-viz-pst2011}.

\subsubsection{Illimitable Space System (ISS)}

Original ISS's research-creation practices focused primarily on interactive multimodal 
installations and productions with the collaboration of local artistic 
troupes. It helped mobilizing traditional artists and makes them aware of the 
new technology in order to express themselves in the new form of artistic 
approach. It started off as a new HCI in the theatre concept and interactive 
documentaries and moved to performing arts and alternate realities. Various 
versions of Illimitable Space System exist for motion capture, signal 
processing, computer vision, projection mapping including LED control, real-time
reaction and control for stage and beyond
\cite{%
msong-phdthesis-2012,%
iss-v3-ar-vr-sa2016,%
multicamtk-siggraph2016,%
iss-v2-design-theory-journal,%
rapid-multimodal-apps-course-siggraph2017%
}.

ISS and its open-source backend core OpenISS~\cite{openiss} rely on computer vision techniques and machine learning
provided by OpenCV and {\marf}; motion capture libraries for Kinect depth cameras and 
others, sound control, input from voice and music, and augmented and virtual 
reality components to co-create either augmented performance or have an 
installation or film, or use as an education tool for artists \cite{rapid-multimodal-apps-course-siggraph2017} or children.

\paragraph{``Projected Reality''}

We explore an idea of a scalable management, visualization,and evaluation of digital evidence in the context 
for cybercrime investigation with extensions to the interactive 3D documentary subsystem of the 
\emph{Illimitable Space System} (ISSv1)~\cite{msong-phdthesis-2012}.
These modifications would enable investigators to represent and create semantic links among digital evidence within an easy to 
use interface powered by multimodal interactions including but not limited to eye-gaze, gestures and navigational hardware.
That work may scale when properly re-engineered and enhanced to act as an interactive ``3D window” into the 
evidential knowledge base grouped into the semantically linked ``bubbles' visually representing the 
documented evidence. By moving such a contextual window, or rather, navigating within the 
theoretically illimitable space an investigator can sort out and reorganize the knowledge items as 
needed prior launching the reasoning computation. The interaction design aspect would be of a 
particular usefulness to open up the documented case knowledge and link the relevant witness accounts 
and group the related knowledge together. This is a proposed solution to the large-scale visualization 
problem of large volumes of ``scrollable'' evidence that does not need to be all visualized at once 
but behave like a snapshot of a storage depot~\cite{mokhov-phd-thesis-2013}.

As an example,
stills from the actual ISSv1 installation hosting multimedia data (documentary videos) users can 
call out by voice or gestures to examine the contents as in \xf{fig:iss-interactive-docu-bubbles}\footnote{http://vimeo.com/51329588}. 
We propose to reorganize the latter into more structured spaces so that the investigators can create semantic 
links to group the relevant evidences together and for subsequent evaluation by the distributed
{\gipsy}'s backend engine~\cite{mokhov-phd-thesis-2013}. As exemplified in ISSv1 the interactions
here are projected on a wall/screen or appear on a monitor.
Currently, the viewable scene/window is sequentially loaded and unloaded from the viewing 
device (PCs, laptops, or VR headsets) to prevent memory overload. The access is on demand 
by the device and the design is similar to RAM swapping by an operating system to support 
virtual memory and particularly in this case a distributed storage with evidential data. 
Available gesture-based interactions using Kinect and similar depth cameras with OpenCV 
are the enabling HCI aspects for the investigator to link the evidential items in the 3D space.
The gesture-based interactions provide optional assistance by 
the voice-based controls for speech processing and the corresponding
commands to view the evidence in detail. Modern availability of 
VR headsets and VR phone applications make this process even more accessible, 
although the storage, space and bandwidth requirements have higher constraints, 
to begin with.
In our general approach, we propose an architecture to enable interactive 
visual windowing into the digital evidence processing as an investigator aid tool. Thus, 
the preferred method of interaction during analysis and human insight phases prior to or after 
distributed processing of the evidence and event reconstruction algorithms.

\begin{figure*}[htpb]
\begin{center}
	\subfigure
	[Corner-projected interactive wall]
	{\includegraphics[width=.47\textwidth]{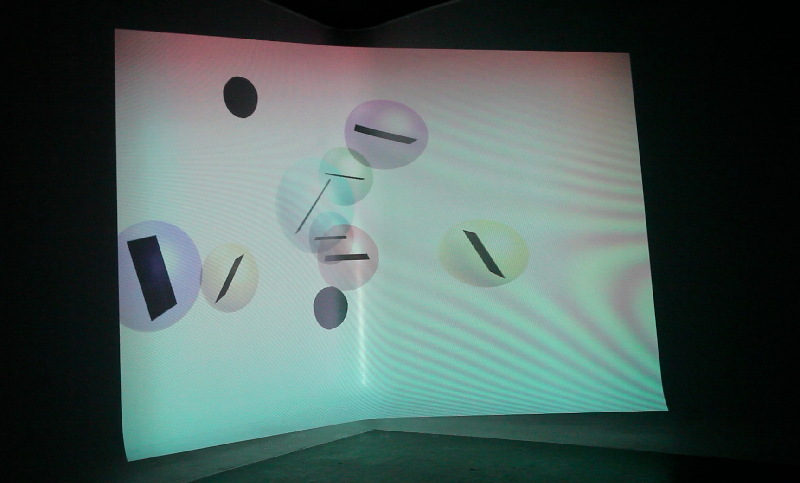}
	 \label{fig:bubble-installation-1}}
	\subfigure
	[Monitor rendering of the interactive 3D window]
	{\includegraphics[width=.47\textwidth]{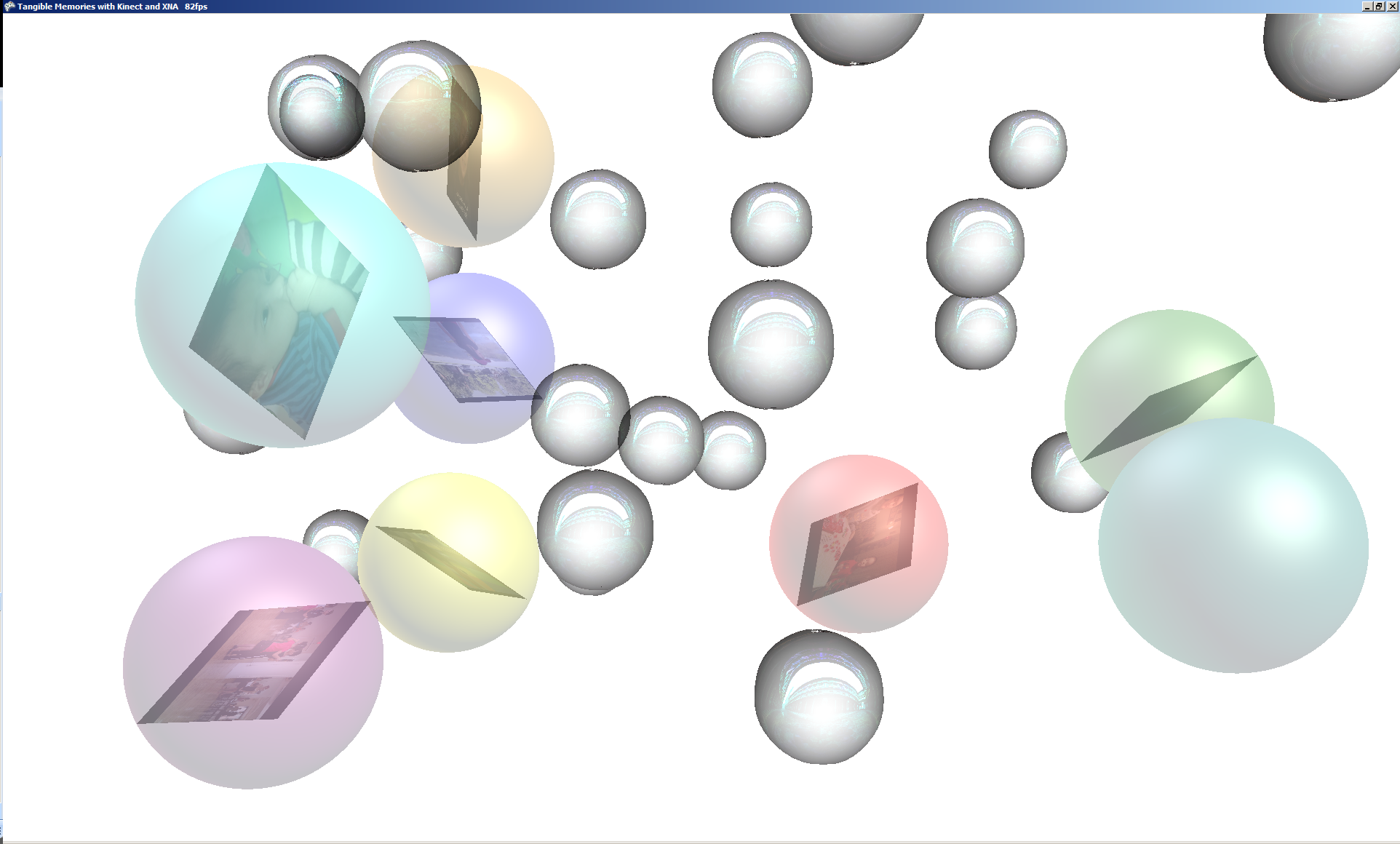}
	 \label{fig:tangible-kinect-2-white}}
	\subfigure
	[Wall-projected interactive wall]
	{\includegraphics[width=.47\textwidth]{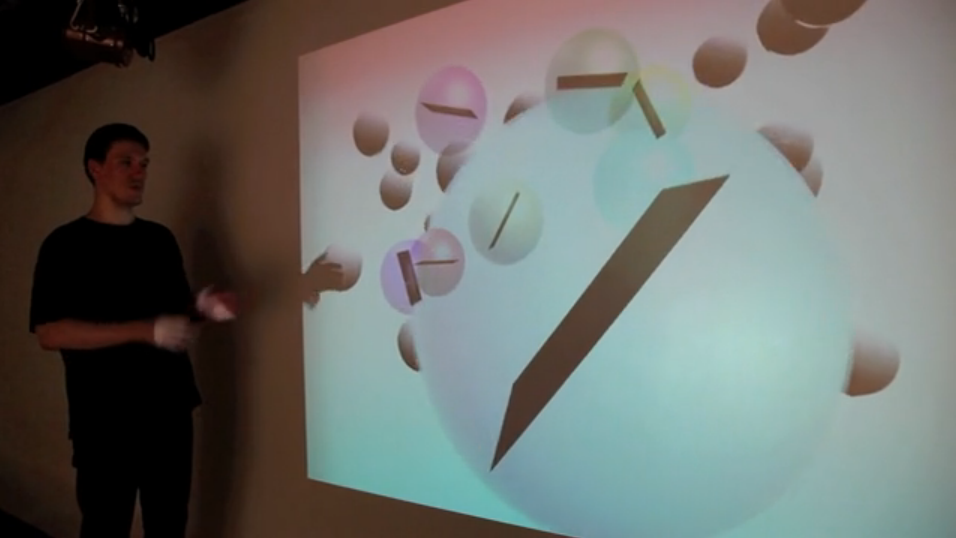}
	 \label{fig:bubble-installation-3}}
	\subfigure
	[Conceptual 3D visualization and rendering (future)]
	{\includegraphics[width=.47\textwidth]{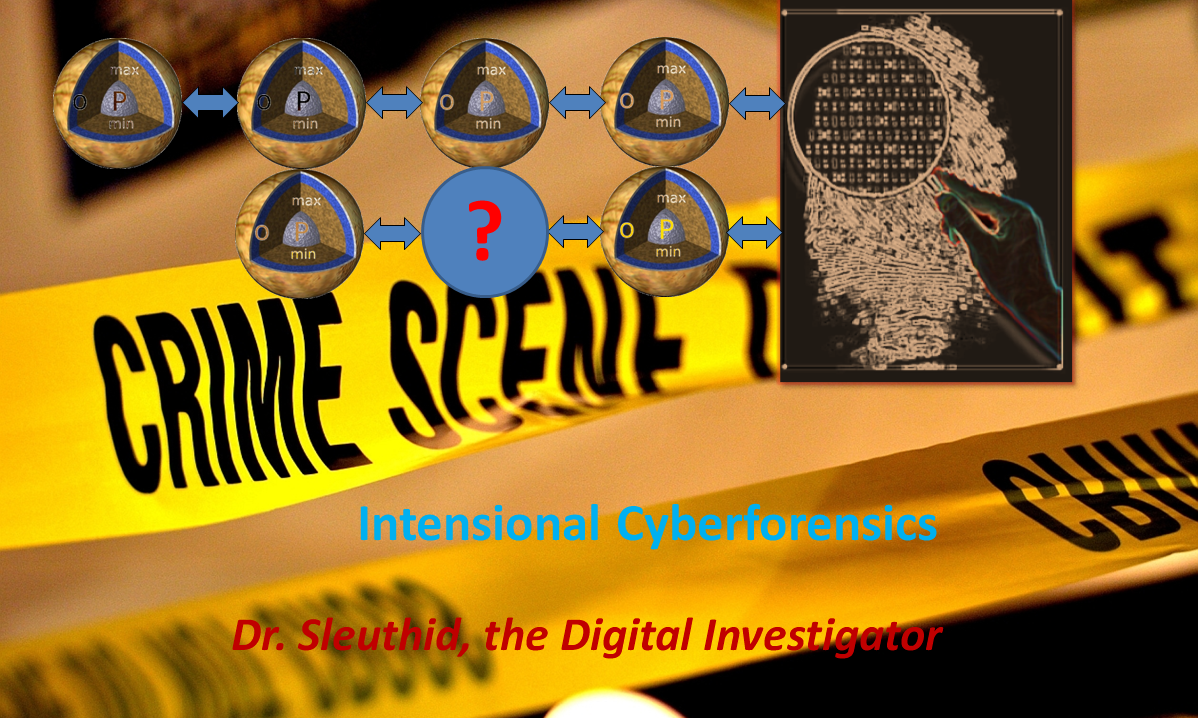}
	 \label{fig:1d-elastic-object}}
\caption{Interactive documentary using Illimitable Space System (ISSv1) visualization and management \cite{msong-phdthesis-2012}}
\label{fig:iss-interactive-docu-bubbles}
\end{center}
\end{figure*}

\paragraph{Virtual and Augmented Reality}

Virtual Reality (VR) \cite{the-vr-book-2016,vr-interactions-course-siggraph2017,course-context-aware-3d-gesture-vr-s2015}
is defined in \cite{mr-taxonomy-1994} as ``a three-dimensional simulation of 
the real world or an imaginary world allowing the user to have a sense of 
physical presence and to manipulate 3D objects,
in real-time, inside three-dimensional computer-generated environments.''
In \cite{vr-educational-tool-1995}, authors point out the possibility of exhibiting concepts that a user might not be able to 
view otherwise and the immersive nature of VR can aid in education thus, can 
be inferred for investigators as well.

Augmented Reality (AR) has to do with overlaying virtual objects on top of the
real ones and a possibility with gesture or gaze based interaction with
these objects while maintaining a grasp on the real world without complete
immersion (avoiding nausea and other related VR issues).

The real benefit is when both techniques can be combined.
ISSv3 in our case was being developing incorporating augmented and virtual
reality techniques~\cite{iss-v3-ar-vr-sa2016,iss-v3-appy-hour-gem2015}.
We experimented in doing both mobile and desktop version of the mixed
reality documentary and recording functionality in Unity some of which
is visualized in \xf{fig:representative_image-smudge} and in \xf{fig:iss-v3-ar-vr-some-more}.

\begin{figure*}%
\includegraphics[width=\textwidth]{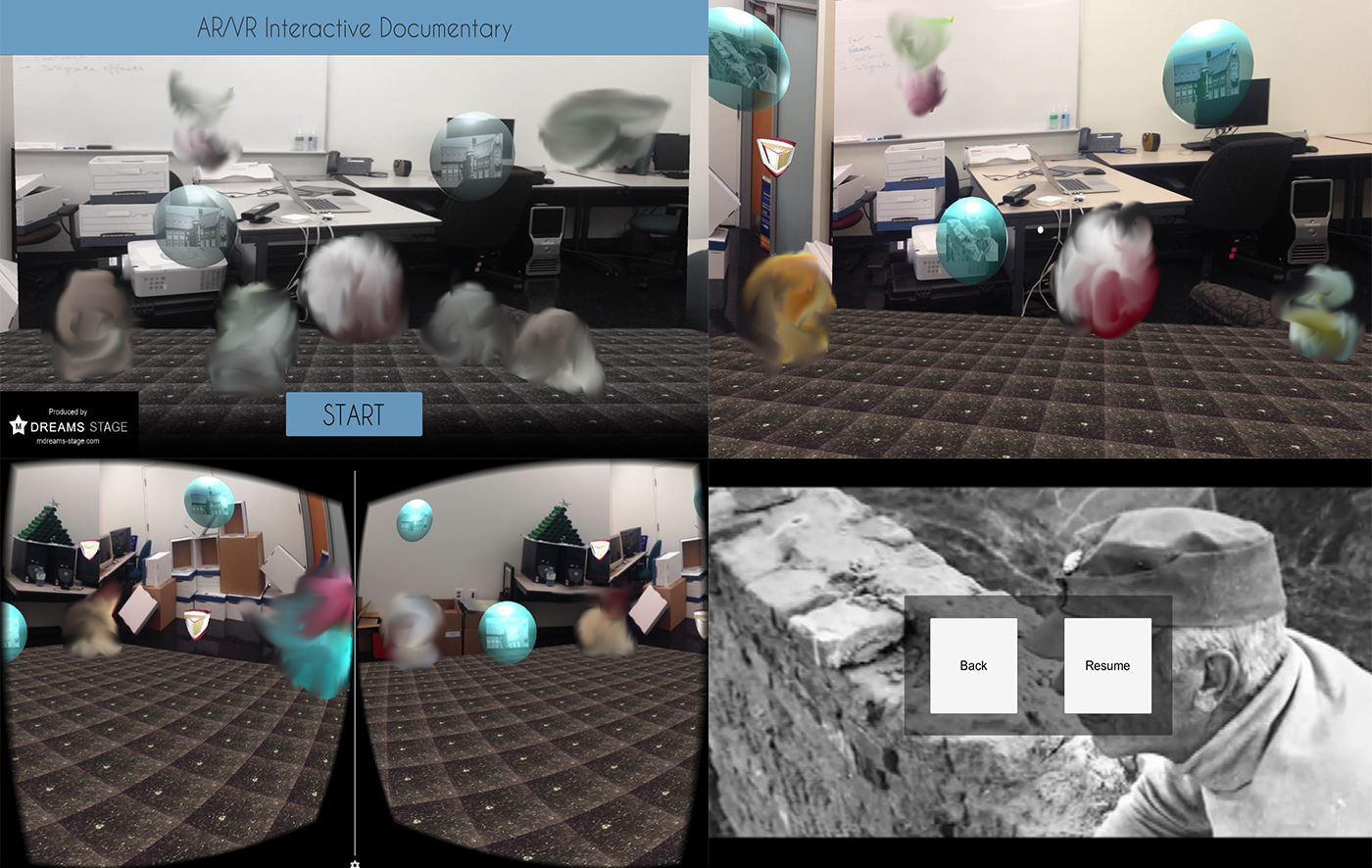}%
\caption{ISSv3 Examples of th VR environment and some digital content \cite{iss-v3-ar-vr-sa2016}}%
\label{fig:representative_image-smudge}%
\end{figure*}

\begin{figure*}[htpb]
\begin{center}
	\subfigure
	[VR version of AR Concordia buildings; interactive]
	{\includegraphics[width=.47\textwidth]{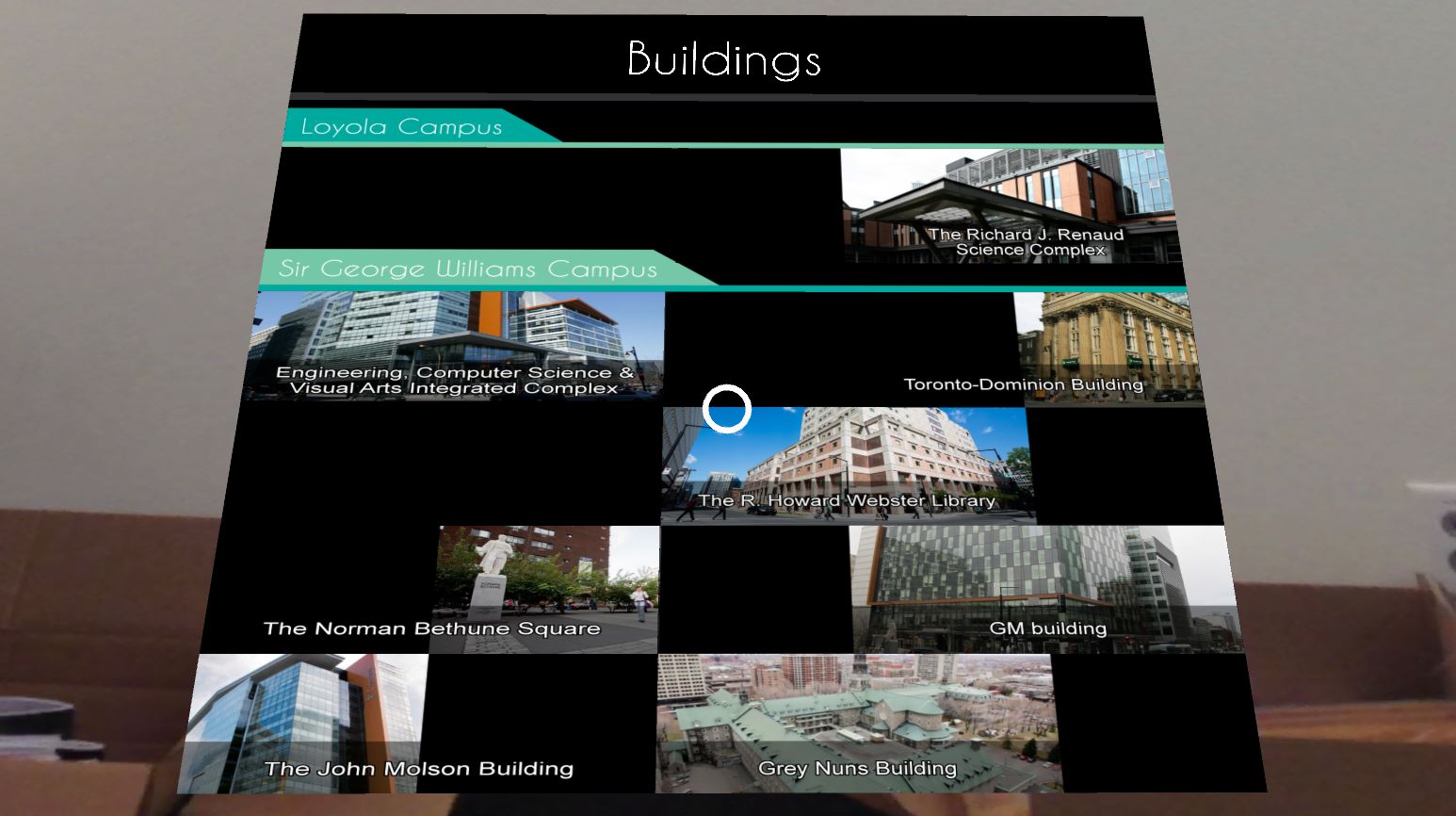}
	 \label{fig:buildings_panel}}
	\subfigure
	[AR camera on for photograph images]
	{\includegraphics[width=.47\textwidth]{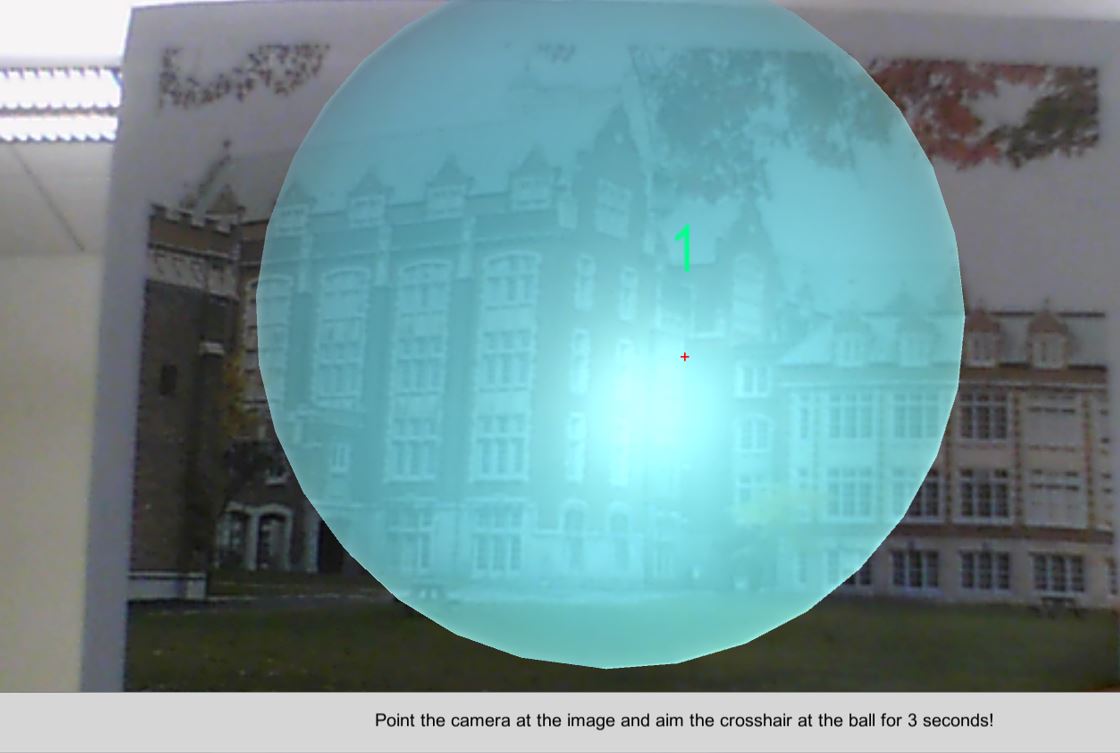}
	 \label{fig:intro_bubble}}
	\subfigure
	[AR/VR lab]
	{\includegraphics[width=.47\textwidth]{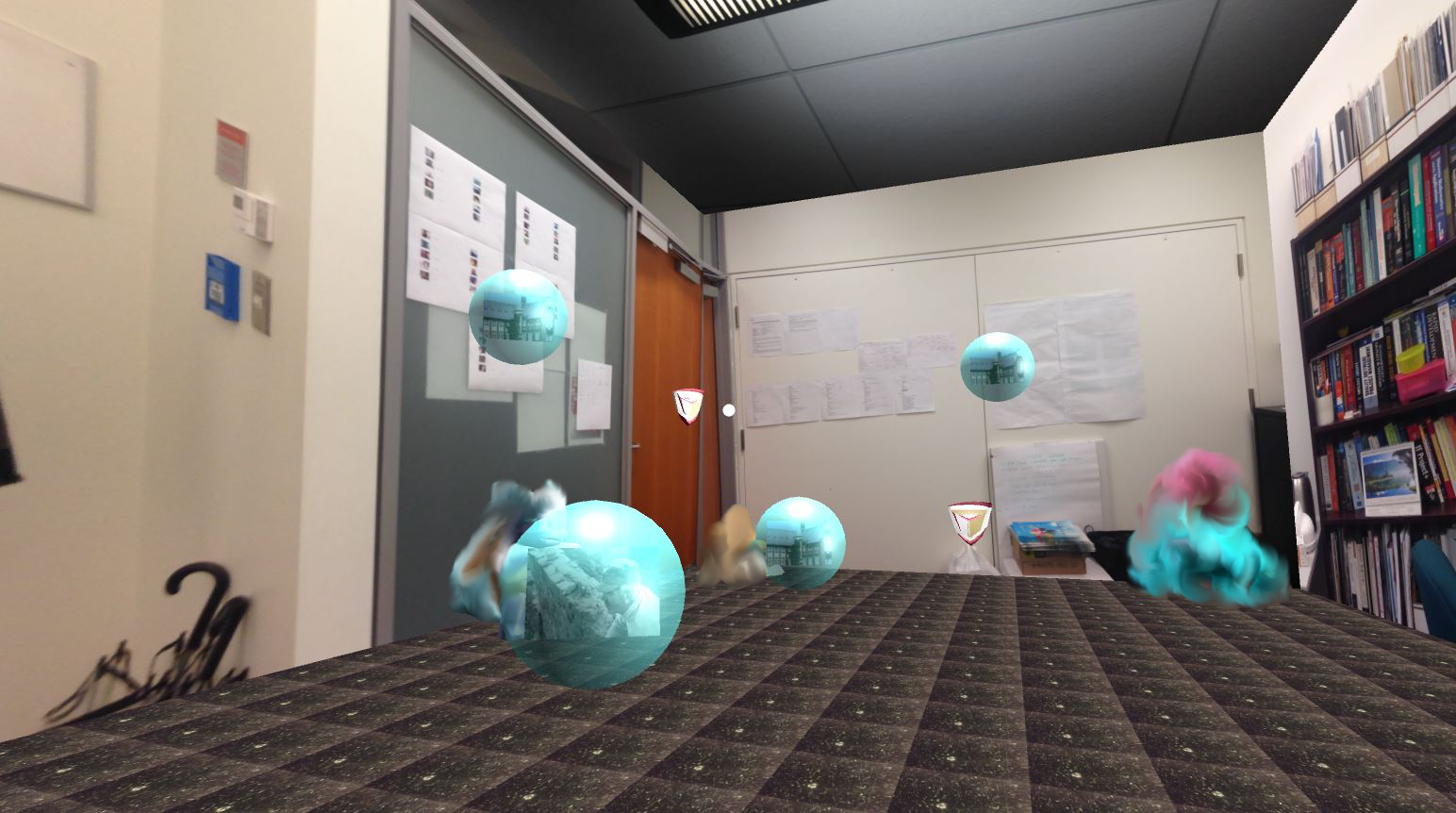}
	 \label{fig:room2-smudge}}
	\subfigure
	[Bubbles with footage in them]
	{\includegraphics[width=.47\textwidth]{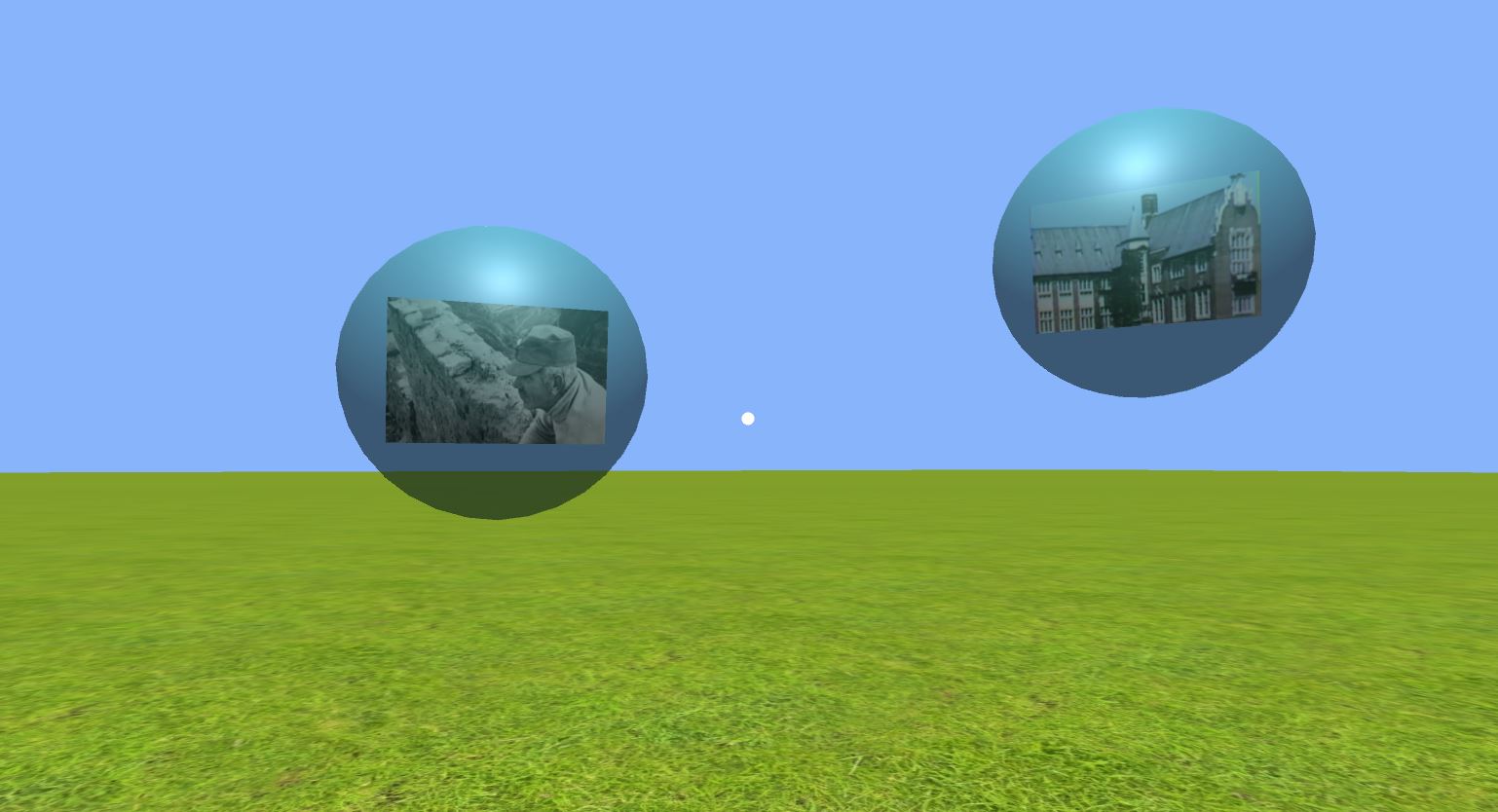}
	 \label{fig:two_bubbles}}
\caption{Interactive documentary using Illimitable Space System (ISSv3) VR/AR \cite{iss-v3-ar-vr-sa2016}}
\label{fig:iss-v3-ar-vr-some-more}
\end{center}
\end{figure*}

\section{Concluding Remarks}

Moving towards our goal to have a visual 3D {\dfg}-based tool that can model 
{\flucid} case specification, and above discussed choices that provide 
the abilities to do the same in their own ways we attempt to build upon 
related research work in this area. However, we do consider the potential of 
the recent work in virtual reality and augmented reality along with 
different multimodal interaction techniques that seem most consistent with 
our aim. So far, Ding's work on {\graphviz}, Puckette's {\puredata}, {\bpel} and the 
ISS have drawn our interest and all of them are sound and formally backed 
standards with some exposure in the industry. While the others may require 
additional work to specify the credibility and correctness of the 
bidirectional translation between 3D DFG visualization and {\flucid}~\cite{flucid-dfg-viz-pst2011}.

The drawbacks of {\puredata} and {\graphviz}s \tool{dot} are that their languages lack 
formal semantics specifications with a few semantic notes along with lexical 
and grammar related structures~\cite{dot-language}. Thus, employing any or all of these will 
require us to provide translation rules and their equivalent semantics to 
{\flucid} as in Jarraya work that provides translations between the UML2.0/SysML
state/activity diagrams and probabilities in~\cite{vv-uml-sysml-syseng-designs}
when translating to {\prism}~\cite{flucid-dfg-viz-pst2011}.
ISS is the most scalable approach that can aggregate all the others, but 
requires significant number of modifications. Given recent advancements
in ISSv2 and ISSv3 referenced above including both AR/VR interactions,
ISS makes this approach even more appealing and feasible than
previously stated~\cite{flucid-dfg-viz-pst2011}.


\bibliographystyle{IEEEtran}
\bibliography{flucid-vis-docu}

\end{document}